\documentclass[runningheads,a4paper]{llncs}
\usepackage[T1]{fontenc}
\usepackage{amssymb}
\usepackage{graphicx}
\usepackage[mathscr]{eucal}
\usepackage{amsmath}
\usepackage{amsfonts}
\def\E{\mathbb{E}}
\def\R{\mathbb{R}}
\def\A{{\bf A}}

\def\G{{\bf G}}

\def\U{{\bf U}}
\def\V{{\bf V}}
\def\I{{\bf I}}
\def\O{{\bf O}}
\def\0{{\bf 0}}
\def\1{{\bf 1}}
\def\T{{\bf T}}
\def\a{{\boldsymbol{a}}}
\def\x{{\boldsymbol{x}}}
\def\y{{\boldsymbol{y}}}

\def\f{{\boldsymbol{f}}}

\def\z{{\boldsymbol{z}}}
\def\w{{\boldsymbol{w}}}
\def\v{{\boldsymbol{v}}}
\def\vbeta{{\boldsymbol{\beta}}}
\def\vvarepsilon{{\boldsymbol{\varepsilon}}}
\def\evbeta{\widehat{\boldsymbol{\beta}}}

\def\vLambda{{\bf{\Lambda}}}
\def\evw{\widehat{\boldsymbol{w}}}
\def\tvw{\widetilde{\boldsymbol{w}}}

\def\ovz{\overline{\boldsymbol{z}}}
\def\ew{\widehat{w}}
\def\tw{\widetilde{w}}

\def\tvbeta{\widetilde{\boldsymbol{\beta}}}

\def\esigma{\widehat{\sigma}}
\def\f{{\boldsymbol{f}}}

\def\ek{\widehat{k}}
\def\oz{\overline{z}}
\def\eR{\widehat{R}}
\def\kmax{k_{\rm max}}
\def\kfolds{k_{\rm folds}}
\def\nunlab{n_{\rm unlab}}
\begin{document}
\title{A semi-supervised learning using over-parameterized regression}
\titlerunning{Over-parameterized regression for semi-supervised learning}
%
\author{Katsuyuki Hagiwara}
\authorrunning{K. Hagiwara}
%
\institute{Faculty of Education, Mie University,\\ 1577
Kurima-Machiya-cho, Tsu, 514-8507, Japan\\
\email{hagi@edu.mie-u.ac.jp}}
\maketitle              
\begin{abstract}
Semi-supervised learning (SSL) is an important theme in machine
learning, in which we have a few labeled samples and many unlabeled
samples. In this paper, for SSL in a regression problem, we consider a
method of incorporating information on unlabeled samples into kernel
functions. As a typical implementation, we employ Gaussian kernels whose
centers are labeled and unlabeled input samples. Since the number of
coefficients is larger than the number of labeled samples in this setting,
this is an over-parameterized regression problem. A ridge
regression is a typical estimation method under this setting. In this
paper, alternatively, we consider to apply the minimum norm least
squares (MNLS), which is known as a helpful tool for understanding deep
learning behavior while it may not be application oriented.  Then, in
applying the MNLS for SSL, we established several methods based on
feature extraction/dimension reduction in the SVD (singular value
decomposition) representation of a Gram type matrix appeared in the
over-parameterized regression problem. The methods are thresholding according to
singular value magnitude with cross validation, hard-thresholding with
cross validation, universal thresholding and bridge thresholding
methods.  The first one is equivalent to a method using a well-known low
rank approximation of a Gram type matrix. We refer to these methods as
SVD regression methods. In the experiments for real data, depending on
datasets, clear superiority of the proposed SVD regression methods over
ridge regression methods was observed. And, depending on datasets,
incorporation of information on unlabeled input samples into kernels was
found to be clearly effective.

\end{abstract}
\section{Introduction}

\subsection{Background, motivation and related works}

Semi-supervised learning (SSL) is an important theme in machine
learning\cite{SSLSurvey,Yang+2023}.  It is a situation where we have a
few input-output samples and many input samples for which we have no
corresponding output samples; e.g., \cite{SSLSurvey}. This problem
setting is mainly studied in image classification, in which the latter
samples are often referred to as unlabeled samples; e.g.,
\cite{Fan+2023}. However, a little have been reported for regression
problems; e.g. \cite{ZL2005,Yang+2023}.  In \cite{ZL2005}, two
$k$-nearest neighbor regressors with different distance metrics are
trained in parallel and used to generate pseudo-labels for unlabeled
data according to a confidence measure; i.e., it employs co-training. In
\cite{Yang+2023}, two reasonable methods have been considered, which are
data augmentation consistency regularization and empirical risk
minimization on augmented data. In both methods, data augmentation is
introduced, in which new input sample is generated by adding noise to an
training input.  Therefore, as shown in \cite{Bishop1995}, these methods
implement certain regularization effects.  In this paper, as a new-type
of SLL for regression problems, we propose a method in which information
on unlabeled inputs is taken into a regressor. More precisely, we set a
kernel type regressor whose components are determined by both of labeled
and unlabeled inputs. For example, in this paper, we employ Gaussian
kernel functions whose centers are labeled and unlabeled input
samples. Since the number of kernels is larger than the number of
labeled training samples, this setting is a so-called over-parameterized
case. Therefore, we need modeling methods under an over-parameterized
case.

A straight way of modeling in this setting is an $\ell_2$ regularization
method which is equivalent to a ridge regression. It solves a
colinearity problem in an over-parameterized case and yields a
reasonable smooth output by optimizing a ridge/regularization
parameter. On the other hand, the minimum norm least squares (MNLS) is
also an estimation procedure for over-parameterized regression problems.
Although it has been of interest in deep learning as a tool for
analyzing ``double decent'' and ``benign
over-fitting''\cite{Belkin+2019,Hastie+2022,Tsigler+2023}, there are no
effective applications of MNLS in regression problems.  Indeed, the
prediction performance of a naive MNLS solution has shown to be worse
than that of a naive ridge regression\cite{Hastie+2022}.  Therefore, we
need a certain modification of MNLS in our proposed
SSL scheme.

In \cite{Dhillon+2013,Tsigler+2023}, it has been shown that a naive
ridge regression is dominated by a type of principal component
regression (PCR). PCR is a method, in which an input data matrix is
transformed into a score matrix and, then, ordinal least squares method
is applied by employing the low-dimension score matrix with large eigen
values as an input data matrix.  This can be viewed as a regression
method under a low rank approximation of input data matrix.  Actually,
\cite{Dhillon+2013} has showed a typical example which prefers PCR. The
MNLS estimate is obtained via singular value decomposition (SVD) of
input data matrix instead of principal component expansion. Then, for
our proposed SSL scheme using MNLS, we consider to implement feature
extraction/dimension reduction in the SVD representation.

\subsection{Contribution}

In this paper, we established SSL scheme based on regression methods in
an over-parameterized case; i.e., over-parameterized regression
methodsw. The idea of the SSL scheme is that we employ a linear
combination of Gaussian kernel functions whose centers are labeled and
unlabeled input samples.  By this implementation, we attempt to
incorporate information on unlabeled inputs into a regressor. 
The choice of Gaussian kernel is not so essential in this idea while it
is a typical choice.  In this setting, the number of kernels is equal to
the number of whole data including labeled and unlabeled samples.
Therefore, this is an over-parameterized regression problem and a
typical estimation method is a ridge regression which is equivalent to
the $\ell_2$ regularization. In this paper, alternatively, we consider
the MNLS estimate obtained via SVD of a Gram type matrix of kernels, in which
a vector of output samples are transformed and a weight vector achieving
perfect fit to the transformed vector is obtained.  The MNLS coefficient estimates are
obtained by an inverse transform of the weight vector.  We refer to this
transformed domain as SVD domain.

By regarding the SVD domain as a feature space, we consider to apply
feature extraction/dimension reduction in the SVD domain and obtain
coefficient estimates by using inverse transform from the modified
weight vector in the SVD domain.  More specifically, we apply
thresholding methods for the weight vector in the SVD domain.  This
strategy is a sparse modeling using a thresholding method on the SVD
domain. Indeed, it generally yields shrinkage estimates of coefficients
in the original domain. We call this strategy SVD regression method. If
we employ a dimension reduction scheme according to the magnitudes of
SVs then it corresponds to PCR in the under-parameterized case; i.e., we
remove components with small PCs/SVs.  Thus, this method implements low
rank approximation of a Gram type matrix in some sense. We refer to this
method as SSV. As pointed out for PCR, this method does not utilize
output samples to choose components. Then, we established thresholding
methods to choose components that are effective for explaining outputs.
We introduced three different methods that are a hard-thresholding (HT)
method using a naive cross validation choice of threshold level, a HT
method with the universal threshold level\cite{DJ1994,DJ1995} and a
bridge thresholding (BT) method that links between soft-thresholding
(ST) and HT\cite{nng,alasso,Hagi2022}.  We refer to these methods as
SHT, SUT and SBT respectively.

We then conducted experiments for real data from UCI
repository\cite{UCI} to examine the effectiveness of this scheme of SSL
and performances of SVD regression methods.  For the latter, we compared
SVD regression methods to ridge regression methods that are referred to
as RR and RRO, RR is a ridge regression under labeled samples and RRO is
a ridge regression under both of labeled and unlabeled samples, thus, in
over-parameterized case. The experimental results are summarized as
follows.

\begin{itemize}
 \item The prediction performances of regression methods depend on
       datasets. There are datasets that prefer ridge regression methods
       and, also, there are datasets that prefer SVD regression
       methods. The difference in the performances is clear when
       the number of training samples are relatively large.

\item Among SVD regression methods, the prediction performances SHT, SUT
       and SBT are comparable to that of SSV which utilizes information
       only about input samples.

\item Depending on the dataset, the
performance of ridge regression is clearly improved by semi-supervised
learning formulated as an over-parameterized case.  Although a similar
result is obtained for SBT, the efficacy may not be so clear.

\end{itemize}

\subsection{Organization of paper}

In Section 2, we explain SSL using over-parameterized regression
methods.  In Section 3, we explain regression problems in an
over-parameterized case, in which we introduce thresholding methods
under the SVD-based MNLS.  The experiments for examining prediction
performances of SSL methods are shown in Section 4.  Section 5 is
denoted for conclusions and future works.

\section{Semi-supervised learning for regression}

\subsection{Problem setting}
\label{sec:problem-setting}

We firstly give several notations. For a matrix $\A$, $\A^\top$ denotes
the transpose of $\A$.  ${\rm diag}(a_1,\ldots,a_m)$ denotes a diagonal
matrix whose diagonal elements are $a_1,\ldots,a_m$.  $\I_n$ and
$\O_{n,m}$ denote the $n\times n$ identity matrix and the $n\times m$
zero matrix respectively. $\0_n$ denotes the $n$-dimensional zero column vector.
For a vector $\a=(a_1,\ldots,a_m)^\top$, we define
$\|\a\|_2=\sqrt{\sum_{i=1}^ma_i^2}$ which is the Euclidean norm of
$\a$. For a real number $u$, we define $(u)_{+}=\max(0,u)$.  If
$\z=(z_1,\ldots,z_m)^{\top}$ has a multivariate Gaussian distribution with mean
vector $\a$ and covariance matrix $\A$, as a usual notation, we write
$\z\sim N(\a,\A)$. This notation is common for one dimensional case.

In a setting of SSL, for $p>n$, we have $n$
input-output data $\{(\x_i,y_i):i=1,\ldots,n\}$ and $p-n$ output data
$\{\x_i:i=n+1,\ldots,p\}$, where $\x_i=(x_{i1},\ldots,x_{id})\in\R^d$
and $y_i\in\R$. The latter is often called unlabeled data in
classification problems and we use this term while our setting is a
regression problem. We assume that
\begin{align}
\label{eq:data-generation}
y_i=f(\x_i)+\varepsilon_i,
\end{align}
where $\varepsilon_1,\ldots,\varepsilon_p$ are i.i.d samples from
$N(0,\sigma^2)$. We define $\y=(y_1,\ldots,y_n)^\top$,
$\f=(f(\x_1),\ldots,f(\x_n))^\top$ and
$\vvarepsilon=(\varepsilon_1,\ldots,\varepsilon_n)^\top$.
Then, (\ref{eq:data-generation}) can be written as
\begin{align}
\label{eq:data-generation-vec}
\y=\f+\vvarepsilon.
\end{align}

\subsection{Proposed scheme}

For $k=1,\ldots,p$, let $g_k$ be a Gaussian function defined by
\begin{align}
\label{eq:Gaussian-kernel}
g_k(\x)=\exp\left(-\frac{1}{\tau}\|\x-\x_k\|_2^2\right),
\end{align}
for $\x\in\R^d$, 
in which the center is the $k$-th input sample and $\tau>0$ is a common
width that is a hyper-parameter. Here, we call $g_1,\ldots,g_p$ kernel
functions while it may not be properly formulated here.

By employing $n$ input-output data as training data, 
we consider to estimate $f$ by
\begin{align}
\label{eq:nonpara-reg-func}
f_{\vbeta}(\x)=\sum_{k=1}^p\beta_kg_k(\x),
\end{align}
where $\vbeta=(\beta_1,\ldots,\beta_p)^\top$ is a coefficient vector.
In this setting, we expect that extra Gaussian kernels whose centers are
unlabeled input samples help prediction on out of training inputs. In
other words, we incorporate unlabeled inputs into a regressor for
utilizing information of unlabeled inputs for prediction.  This is
regarded as a new-type SSL for regression problems. Since $n<p$ holds,
this setting is an over-parameterized case. Therefore, the colinearity
problem arises in the least squares method and we need the other
estimation methods.  We refer this estimation problem to as an
over-parameterized regression problem. Note that the choice of $g_k$
here may not be essential and it is important that $g_k$ should be a
function that incorporates information on unlabeled inputs into modeling. In this
meaning, a Gaussian function for $g_k$ seems to be one good possible choice
for our purpose. In the next section, we show several estimation methods
for an over-parameterized regression problem arising in SSL.

\section{Over-parameterized regression methods}

We define $g_{i,k}=g_k(\x_i)$.  Let $\G$ be an $n\times p$ matrix whose
$(i,k)$ entry is $g_{i,k}$. $\G$ is a Gram type matrix; i.e., 
it may not be a usual $n\times n$ Gram matrix. We assume that $\G$ is
fixed. We define
$\f_{\vbeta}=(f_{\vbeta}(\x_1),\ldots,f_{\vbeta}(\x_p))^\top:=\G\vbeta$. We
assume that the rank of $\G$ is $n$.

\subsection{Ridge regression}
\label{sec:ridge-regression}

A typical method for over-parameterized regression problems is a ridge
regression which solves the colinearity problem.
In a ridge regression, estimate of a coefficient vector is obtained by
\begin{align}
\label{eq:ridge-estimator}
\evbeta_{\lambda}=(\G^\top\G+\lambda\I_p)^{-1}\G^\top\y,
\end{align}
where $\lambda>0$ is called a ridge parameter.  $\evbeta_{\lambda}$ is
also a solution in minimizing the $\ell_2$ regularized cost function
given by
\begin{align}
C_2(\vbeta):=\|\y-\G\vbeta\|_2^2+\lambda\|\vbeta\|_2^2.
\end{align}
In this formulation, the ridge parameter is called a regularization
parameter.  This method is possible to apply even when $n\le p$.  The
ridge (regularization) parameter is a hyper-parameter which may
be selected by cross validation (CV) typically.
We refer to a naive ridge regression in case of $p\le n$ as RR and 
a ridge regression in $p>n$ (over-parameterized case) as RRO.

As alternatives to a ridge regression method for an
over-parameterized case, we now consider methods based on the MNLS.

\subsection{Minimum norm least squares}

In deep learning, the MNLS estimation has been extensively studied as a
tool for analyzing ``double decent'' and ``benign
over-fitting''\cite{Belkin+2019,Hastie+2022,Tsigler+2023}. The MNLS is a
method obtaining a unique solution of perfect fit in an
over-parameterized case and the MNLS solution is obtained via
SVD. Although the MNLS attracts attention due to the theoretical
interest in deep learning, there is no application to regression
problems.  In applying MNLS for the proposed SSL scheme, we consider to
modify the SVD based formulation of the MNLS. To do this, we briefly
review the MNLS estimation in this subsection.

The MNLS solution is given by minimizing
$\|\vbeta\|_2$ among solutions satisfying
\begin{align}
S(\vbeta):=\|\y-\G\vbeta\|_2^2=0,
\end{align}
therefore, $\y=\G\vbeta$.
By applying SVD to $\G$, we have
\begin{align}
\G=\U\vLambda\V^\top, 
\end{align}
where $\U$ is $n\times n$ orthonormal matrix, $\V$ is $p\times p$
orthonormal matrix and $\vLambda=[\vLambda_1~\O_{n,p-n}]$ in which
$\vLambda_1={\rm
diag}(\lambda_1,\ldots,\lambda_n)$. $\lambda_1,\ldots,\lambda_n$ 
and $\lambda_k>0$ for any $k$ since the rank of $\G$ is
$n$. $\lambda_1,\ldots,\lambda_n$ are called singular values (SVs). 
Without loss of
generality, we assume that $\lambda_1\ge\cdots\ge\lambda_n$ for SVs.
We define
\begin{align}
\label{eq:def-z}
\z&=(z_1,\ldots,z_n)^\top:=\U^\top\y\\
\label{eq:def-w}
\w&=(w_1,\ldots,w_p)^\top:=\V^\top\vbeta.  
\end{align}
Since $\U$ is orthonormal, we
have
\begin{align}
\label{eq:S-beta-w}
S(\vbeta(\w))
&=\|\y-\U\vLambda\V^\top\vbeta\|_2^2\notag\\
&=\|\z-\vLambda_1\w\|_2^2
=\sum_{i=1}^n(z_i-\lambda_iw_i)^2.
\end{align}
Therefore, $S(\vbeta(\w))=0$ for
$\w=\evw:=[\evw_1^\top~\0_{p-n}^\top]^\top$, where
\begin{align}
\evw_1=(\ew_1,\ldots,\ew_n)^\top:=\vLambda_1^{-1}\z. 
\end{align}
Hence, $\ew_k=z_k/\lambda_k$ componentwisely for $k=1,\ldots,n$.  As a
result, the MNLS solution of $\vbeta$ is given by
\begin{align}
\evbeta=\V\evw
\end{align}
since $\V$ is orthonormal. It is easy to see that $\evbeta$ achieves the
minimum norm. Therefore, it is the solution to the MNLS estimation.  In
this derivation, an original $(\G,\y)$ domain is transformed into an
another domain $(\vLambda_1,\z)$ and the fitting is considered in the
latter domain which is determined by singular values.  We refer to the
transformed domain defined by $(\vLambda_1,\z)$ as SVD domain.

We briefly summarize statistical properties of estimates in SVD domain.
We define
$\ovz=(\oz_1,\ldots,\oz_n)^{\top}:=\U\f$.
By (\ref{eq:data-generation-vec}), 
(\ref{eq:def-z}) and orthonormality of $\U$, we have
\begin{align}
\label{eq:E_z}
\E[\z]&=\E[\U\y]=\U\f=\ovz\\
\label{eq:V_z}
\E[(\z-\ovz)(\z-\ovz)^\top]&=\E[\U\vvarepsilon\vvarepsilon^\top\U^\top]
=\sigma^2\I_n.
\end{align}
Hence, $\z\sim N(\ovz,\sigma^2\I_n)$ holds under a Gaussian noise
assumption; i.e., $z_1,\ldots,z_n$ are mutually independent and $z_k\sim
N(\oz_k,\sigma^2)$.  Thus, $\evw_1\sim
N(\vLambda_1^{-1}\ovz,\sigma^2\vLambda_1^{-2})$ holds; i.e.,
$\ew_1,\ldots,\ew_n$ are mutually independent and $\ew_k\sim
N(\oz_k/\lambda_k,\sigma^2/\lambda_k^2)$.
Due to these statistical properties, the $k$-th SV components is
significant if $\oz_k\neq 0$. Therefore, in the SVD regression problem,
we can apply thresholding techniques in
\cite{DJ1994,DJ1995,Hagi2022}.

\subsection{Thresholding in SVD domain}

By (\ref{eq:S-beta-w}), the MNLS solution reduces to a perfect fitting
of $\z$ by $\vLambda_1\w_1$. In other words, $\vLambda_1$ and $\w_1$ are
regarded as a design matrix and coefficient vector respectively; i.e.,
thus, it is an orthogonal regression problem.  We now consider
thresholding on $\evw_1$, which works as forming a sparse representation
for $\z=\U\f$ in terms of $\vLambda_1\w_1$ in the SVD domain.  Note
that, in a usual setting of a sparse modeling\cite{lasso}, a sparseness
is assumed in representing $\f$ by $\G\vbeta$, in which some elements of
$\vbeta$ are zeros in representing $\f$.  On the other hand, we assume
that some elements of $\w$ are zeros in representing
$\z=\U\f$. Therefore, a sparseness is considered in a space that is
transformed by $\U$.  It does not lead to a sparse representation in the
original domain with $\vbeta$ and it rather works as shrinkage of
$\vbeta$. This is because removing components in the SVD domain affects
the entire components in the original domain and the transformation is
isometric; e.g. see \cite{Hagi2011} and later implementations.

Let $t_k$ be a thresholding function on $\ew_k$ for
$k=1,\ldots,n$. $t_k$ may also have a shrinkage effect in a certain method.  We
define
\begin{align}
\label{eq:def-tvw1}
\tvw_1:=(\tw_1,\ldots,\tw_n)^\top=\T(\evw_1),
\end{align}
where $\T(\evw_1)=(t_1(\ew_1),\ldots,t_n(\ew_n))^\top$.
$t_k(\ew_k)=0$ for some $k$ to implement thresholding.
Then, the resulting estimate of $\vbeta$ is obtained by
\begin{align}
\label{eq:def-tvbeta}
\tvbeta=\V\tvw,
\end{align}
where $\tvw=[\tvw_1^\top~\0_{p-n}^\top]^\top$. 
We refer to this estimation scheme as SVD regression. 
In the next section, we show several implementations of 
thresholding in the SVD regression scheme.

\subsection{SV based modeling (SSV)}

The first method is to keep components with large SVs. 
Under a pre-determined
threshold level denoted by $\theta>0$, we set
\begin{align}
\tw_k=\begin{cases}
\ew_k=z_k/\lambda_k & \lambda_k\ge\theta\\       
0 & \lambda_k<\theta.
      \end{cases} 
\end{align}
Let $\v_k$ be the $k$-th column vector of $\V$.  By (\ref{eq:def-w}),
$w_k=0$ is equivalent to $\v_k^{\top}\vbeta=0$ and, thus, 
$\lambda_k\v_k^{\top}\vbeta=0$ in (\ref{eq:S-beta-w}). Therefore, 
this is equivalent to a method of setting $\lambda_k=0$, which is
a low rank approximation of $\G$ in the MNLS estimate.

Here, $\|\evw_1\|^2\ge\|\tvw_1\|^2$ holds. Therefore, we have
\begin{align}
\|\evbeta\|^2&=\|\V\evw\|^2=\|\evw\|^2=\|\evw_1\|^2\notag\\
&\ge\|\tvw_1\|^2=\|\tvw\|^2=\|\V\tvw\|^2=\|\tvbeta\|^2
\end{align}
since $\V$ is orthonormal. This implies that the thresholding in the SVD
domain leads to a shrinkage on the original domain. 
This also applies any other thresholding methods.

We define $K_{\theta}:=\{k:\lambda_k\ge\theta\}$, which is often called
an active set that is a set of indexes of unremoved components.  We
choose $\theta$ in $\theta\in\{\lambda_1,\ldots,\lambda_n\}$, by which
we ignore the case where $K_{\theta}$ is empty; i.e., inclusion of a
constant zero function. Since $\lambda_1\ge\cdots\ge\lambda_n$, if we
set $\theta=\lambda_j$ then $j=|K_{\theta}|$ is the number of unremoved
components, where $|K_{\theta}|$ denotes the number of members in
$K_{\theta}$.  Therefore, the determination of $\theta$ is equivalent to
that of the number of unremoved components. This method is referred to
as SSV in this paper.  Generally, important components for
prediction do not necessarily have large SVs; i.e., SVs do not have
information on a target function.  Therefore, SSV is possible to include
components which are useless for representing a target function while
those have large SVs. Nevertheless, this method may attractive since it
may give us entirely smooth outputs and components with large SVs are
relatively reliable; i.e., include less numerical error.

We consider to apply cross validation (CV) for selecting the number of
unremoved components.  More precisely, in a choice of training and
validation datasets from whole data, SVD is applied to the training data
and validation error is calculated under $k$ unremoved (non-zero)
components for $k=0,1,\ldots,\kmax$. After repeating this procedure for
all splits of whole data, we choose the number of unremoved components,
which minimizes the validation error sum. It is denoted by $\ek$.  Note
that $\kmax$ may be taken to be less than the number of training data
since the SV is sufficiently small and is not reliable for large
$k$. Then, in constructing a final model, after we obtained
$\ew_1,\ldots,\ew_n$ for whole data, we set $\tw_k=0$ forf $k>\ek$ and
$\tw_k=\ew_k$ for $k\le\ek$.

\subsection{Hard-thresholding with CV (SHT)}

We firstly consider a HT, in which a threshold level is selected by CV.
For $\theta_k>0$, 
\begin{align}
\label{eq:ht-func}
\tw_k=H_{\theta_k}(\ew_k):=
\begin{cases}
\ew_k & |\ew_k|\ge\theta_k\\
0 & |\ew_k|<\theta_k 
\end{cases},
\end{align}
is a hard-thresholding (HT) function, where $\theta_k$ is a threshold level
that is a hyper-parameter\cite{DJ1994,DJ1995}.  
Here, $\theta_k$ is set to be different for each $k$. This is because
the variance of $\ew_k$ depends on $k$ by the consequences from
(\ref{eq:E_z}) and (\ref{eq:V_z}).  If we set a common threshold level
for all $\ew_k$ then components with large variances tend to be
unremoved comparing to components with small variances when those means
are equally zeros. In general, the choice of component-wise threshold
levels may be extremely difficult. However, it is not a problem in this
case as follows. We define
$K_{\theta}:=\{k:|z_k|\ge\theta\}$ for $\theta>0$ and set
$\theta_k=\theta/\lambda_k$ for $k=1,\ldots,n$. Then, by
$\ew_k=z_k/\lambda_k$, we have
$K_{\theta}=\{k:|\ew_k|\ge\theta_k\}$. Thus, (\ref{eq:ht-func}) is
equivalent to
\begin{align}
\label{eq:ht-func-z}
\tw_k=
\begin{cases}
\ew_k=z_k/\lambda_k & |z_k|\ge\theta\\
0 & |z_k|<\theta 
\end{cases}
\end{align}
and the active set in this case is $K_{\theta}$.

Now, since $\z\sim N(\ovz,\sigma^2\I_n)$, employing $\theta$ as a common
threshold level on $z_1,\ldots,z_n$ is reasonable. We choose $\theta$
from $\{|z_1|,\ldots,|z_n|\}$ by ignoring the case where $K_{\theta}$ is
empty. Therefore, the selection of $\theta$ is equivalent to the
selection of the number of unremoved components, which is the size of
$K_{\theta}$.  In constructing a regressor for a given data with size
$n$, we obtain $z_1,\ldots,z_n$ and $|z_{j_1}|\ge\cdots\ge|z_{j_n}|$ by
descending enumeration.  Then, we can obtain $\tvw$ with $k$ unremoved
components by setting $\theta=|z_{j_k}|$ in (\ref{eq:ht-func-z}).  One
simple strategy is to determine an optimal number of unremoved
components is to apply CV as in case of SSV. The HT method with CV
choice of a threshold level is referred to as SHT.

\subsection{Universal Thresholding (SUT)}

Again, since $\z\sim N(\ovz,\sigma^2\I_n)$, we can apply the well-known
universal threshold (UT) level in \cite{DJ1994,DJ1995} for
$z_1,\ldots,z_n$. It is given by
\begin{align}
\label{eq:u_threshold_level}
\theta=\sqrt{2\sigma^2\log n}
\end{align}
and is known to be asymptotically
optimal in some sense\cite{DJ1994,DJ1995}. 
We set this $\theta$ in (\ref{eq:ht-func-z}) to obtain $\tvw$.

To apply the UT level, we need an estimate of noise variance denoted by
$\sigma^2$. In wavelet, for example, median absolute deviation (MAD) of
the first detail coefficients is employed\cite{DJ1994,DJ1995}. However,
it cannot be applied to our setting. For estimating noise variance in a
context of general non-parametric regression problems, there are model-based
(residual-based) and model-free (difference-based)
methods\cite{KP+2005,TongWang2005}, in which the former utilizes a ridge
estimate.  There are several variations of residual-based type and we
employ an estimate suggested by \cite{CE1992}.  In our context, in which
$\z$ is fitted by $\vLambda\w$ in the SVD domain, it is given by
\begin{align}
\label{eq:esigma2}
\esigma^2=\frac{\sum_{i=1}^n(1-h_i)^2z_i^2}
{\sum_{i=1}^n(1-h_i)^2},
\end{align}
where $h_i=\lambda_i/(\lambda_i+\lambda)$ and $\lambda>0$ is a ridge
parameter. The prediction performance of the SVD regression method is
determined by thresholding here.  The above ridge parameter is required
for a stable estimation of noise variance and is set to be a small
value; i.e., it is not hyper-parameter in SVD regression. Actually, we
set it to a fixed small value in the later experiments. The HT 
using the UT level is referred to as SUT.

\subsection{Bridge thresholding (SBT)}

As an alternative to HT, soft-thresholding (ST) is
known\cite{DJ1994,DJ1995}.  Unlike HT, ST simultaneously controls both
of threshold level and amount of shrinkage by one hyper-parameter.  It
is well known that ST is a special case of lasso\cite{lasso}; i.e., in
case of orthogonal regression problems. Lasso is known to be suffered
from an estimation bias problem; i.e., choice of a high threshold level
brings us a sparseness while it automatically increases the amount of
shrinkage, which leads to a large bias. For relaxing this problem,
several methods including adaptive lasso\cite{alasso} has been
proposed. In case of orthogonal regression problems, adaptive lasso
reduces to
\begin{align}
\label{eq:al-func}
B_{\theta_k,\gamma}(w):=\left(1-
\left(\frac{\theta_k}{|w|}\right)^{1+\gamma}\right)_{+}w, 
\end{align}
where $\theta_k>0$ and
we set $\gamma\in\{1,3,5,\cdots\}$ for simplicity\cite{Hagi2022}.  Formally, 
$B_{\theta_k,\gamma}$ reduces to ST when $\gamma=0$.  When $\gamma=1$,
it reduces to non-negative garrote under an orthogonal regression
problem\cite{nng}.  On the other hand, if $\gamma$ is very large then
$B_{\theta_k,\gamma}(w)\simeq w$ holds for $|w|>\theta_k$ even when
$|w-\theta_k|$ is small.  Therefore, $B_{\theta_k,\gamma}$ is close to the
HT function as $\gamma\to\infty$; e.g. \cite{Hagi2022}. This implies
that $B_{\theta_k,\gamma}$ yields a bridge estimator that connects
between ST and HT estimators. $B_{\theta,\gamma}$ is also
obtained by scaling a ST estimator to move it to a HT
estimator\cite{Hagi2022}. Therefore, by choosing a relatively large
$\gamma$, the above estimation bias problem of ST is
solved in $B_{\theta,\gamma}$. We refer to this
thresholding method as a bridge-thresholding (BT) method.
In case of applying BT to SVD regression, we have
\begin{align}
\label{eq:bt-func}
\tw_k=B_{\theta_k,\gamma}(\ew_k)
\end{align}
for $k=1,\ldots,n$. As in UT, if we set $\theta_k=\theta/\lambda_k$ 
in (\ref{eq:bt-func}) then we have
\begin{align}
\label{eq:bt-func-z}
\tw_k
&=\begin{cases}
\left(1-\left(\theta/z_k\right)^{1+\gamma}\right)z_k/\lambda_k& 
|z_k|>\theta\\
0 & |z_k|\le\theta
  \end{cases}
\end{align}
since $\gamma$ is odd and $\ew_k=z_k/\lambda_k$.  Therefore, as in HT,
it is enough to choose an appropriate value for $\theta$ instead of
$\theta_1,\ldots,\theta_n$ independently.

Since we set a relatively large value for $\gamma$ to relax bias
problem of ST, $\theta$ is only a hyper-parameter.
Unlike HT, we can derive a risk estimate under BT
\cite{Tibshirani2015,Hagi2022}. 
We define the risk by
\begin{align}
\label{eq:R-theta-0}
R(\theta)&:=\frac{1}{n}\E_{\z}\left[\|\ovz-\vLambda_1\tvw_1\|_2^2\right],
\end{align}
where $\E_{\z}$ denotes the expectation with
respect to the joint distribution of $\z$; i.e., $N(\ovz,\sigma^2\I_n)$
by (\ref{eq:E_z}). 
We define an active set by
$K_{\theta}:=\{k:|z_k|>\theta\}$
and $\ek_{\theta}=|K_{\theta}|$. Note that it is equivalent to
$K_{\theta}=\{k:|\ew_k|>\theta_k\}$ as in HT.
In \cite{Tibshirani2015,Hagi2022}, by using Stein's lemma\cite{stein}, 
\begin{align}
\label{eq:bt-criterion}
\eR(\theta)=&\frac{1}{n}\sum_{i=1}^n
(z_i-\lambda_i\tw_i)^2-\sigma^2
+\frac{2\sigma^2}{n}
\left(\ek_{\theta}+\gamma\sum_{i\in K_{\theta}}
\left(\frac{\theta}{z_i}\right)^{\gamma+1}\right).
\end{align}
has been found to be an unbiased estimate of the risk $R(\theta)$, which
is called Stein's unbiased risk estimate (SURE). This can be a model
selection criterion for $\theta$; i.e., calculate $\eR(\theta)$ for
$\theta\in\Theta$ and choose $\arg\min_{\theta\in\Theta}\eR(\theta)$,
where $\Theta$ is a candidate set of $\theta$. In applications, as in
SUT, we need to estimate $\sigma^2$ and employ (\ref{eq:esigma2}) for it
again. This method is referred to as SBT.  In SBT, although $\theta$
determines not only the threshold level but also the amount of
shrinkage, we choose a value of $\theta$ from $\{|z_1|,\ldots,|z_n|\}$ in the experiments
for simplicity.

\section{Numerical examples}

In this section, we consider a situation where we have $n$ input-output
training samples and $\nunlab$ unlabeled input samples.  As alternatives to
the SVD regression methods, we consider ridge regression methods with
$p=n$ and $p>n$ in the subsection \ref{sec:ridge-regression} under
(\ref{eq:Gaussian-kernel}). Those are referred to as RR and RRO
respectively; i.e., RR does not utilize un-labeled inputs and RRO
does. In a context of semi-supervised learning, we compare the
performances of RR, RRO, SSV, SHT,
SUT and SBT.

In this paper, we show results on four typical example datasets from UCI machine
learning repository\cite{UCI}, which are
\begin{itemize}
\item SGEMM GPU kernel performance dataset : the number of whole data is
      $10000$ (a part of available data), the number of input variables
      (features) is $14$ and a target variable is ``Run2'',
\item Auction Verification dataset : the number of whole data is $2043$, the number of
      features is $7$ and a target variable is ``verification.time'',
\item Energy efficiency dataset : the number of whole data is $768$, the number of
      features is $8$ and a target variable is ``Y1'',
\item Yacht Hydrodynamics dataset : the number of whole data is $308$, the number of
      features is $6$ and a target variable is ``resistance''; 
\end{itemize}
see \cite{UCI} for details of variables. The features are 
normalized in our experiments.

The setting of hyper-parameters in the methods is as follows.  The
hyper-parameters which are selected by cross validation are the ridge
parameter in (\ref{eq:ridge-estimator}) and width parameter in
(\ref{eq:Gaussian-kernel}) for RR and RRO, the number of components and
width parameter for SSV and SHT, the width parameter for SUT and SBT.
The number of folds of cross validation is $\kfolds=10$.  In RR and RRO,
the candidate values of hyper-parameters are $\{n\times 10^{-q}:
q=0,2,\ldots,16\}$ for a ridge parameter and
$\{10^{q}:q=-1,0,1,\ldots,6\}$ for a width parameter, in which $n$ is
the number of training data.  In SVD regression methods, the candidate
values of hyper-parameters are $\{10^{-q}: q=-1,0,1,\ldots,10\}$ for a
width parameter and $\{0,\ldots,\kmax\}$ for the number of unremoved
components. Here, $\kmax$ is set to the integer part of $2n/3$. The SVs
in this range may be stable, in which $\kmax$ may be sufficiently
large. Note that, for almost experiments, values of hyper-parameters are
within the above candidates.  The variance estimate is obtained by
(\ref{eq:esigma2}) in SUT and SBT, in which we set $\lambda=10^{-12}$
which stabilizes the estimate. In SBT, the number of components is
selected by (\ref{eq:bt-criterion}) with this variance estimate and we
set $\gamma=7$ which is enough for relaxing a bias problem of ST.

In the first experiment, we compare the performances of the methods.  In
Fig.\ref{fig:methods-comparison}, we show the mean and standard
deviation (error bar) of test errors of the methods for $50$ trials with
different random choices of training and test sets, in which test error
is measured by $1-R^2$; i.e. ratio of the mean squared test error of
estimate to that of the mean of test outputs. For any $n=50,100,200$, we
set $\nunlab=100$. As seen in Fig.\ref{fig:methods-comparison} (a) and
(b), for GPU and Auction datasets, ridge regression methods are superior
to SVD regression methods. This is reliable and notable in case of
$n=200$.  This implies that, for these datasets, SVD may not be
appropriate as feature selection for helping prediction. Among
SVD regression methods, SSV is relatively superior to the other
methods. This implies that components with large SVs contributes stably
for prediction and choice of contributed components according to output
data does not work so well for these datasets.  Note that the results
for $n=50$ in Fig.\ref{fig:methods-comparison} (a) and (b) may not be
reliable in all methods since their standard deviations are large, which
implies that $n=50$ may not be sufficient for prediction.  And, as seen
in case of $n=200$, any method can stably produces better estimate when
$n$ is sufficiently large.

\newcommand{\figwidth}{45mm}

\begin{figure}[h]
\begin{center}

\begin{minipage}[t]{\figwidth}
\begin{center}
\includegraphics[width=\figwidth]{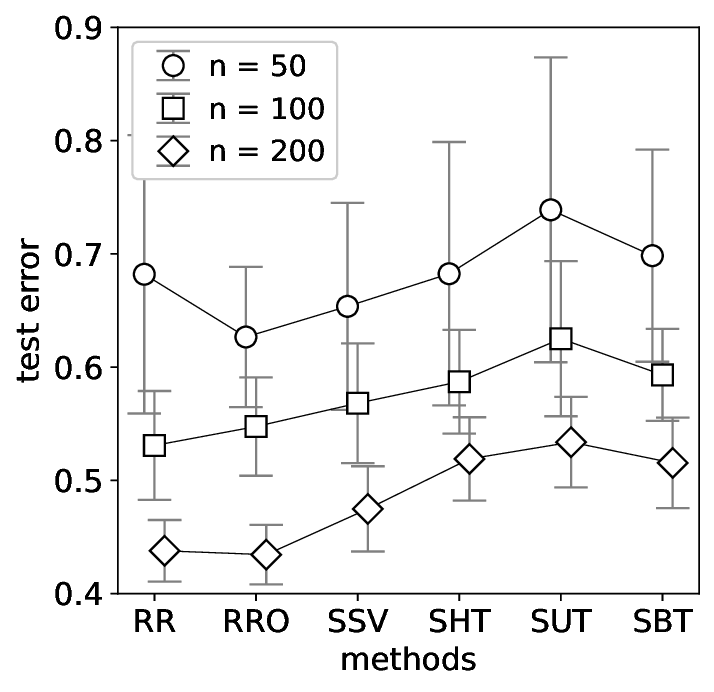}

(a)  GPU dataset
\end{center}
\end{minipage}
\begin{minipage}[t]{\figwidth}
\begin{center}
 \includegraphics[width=\figwidth]{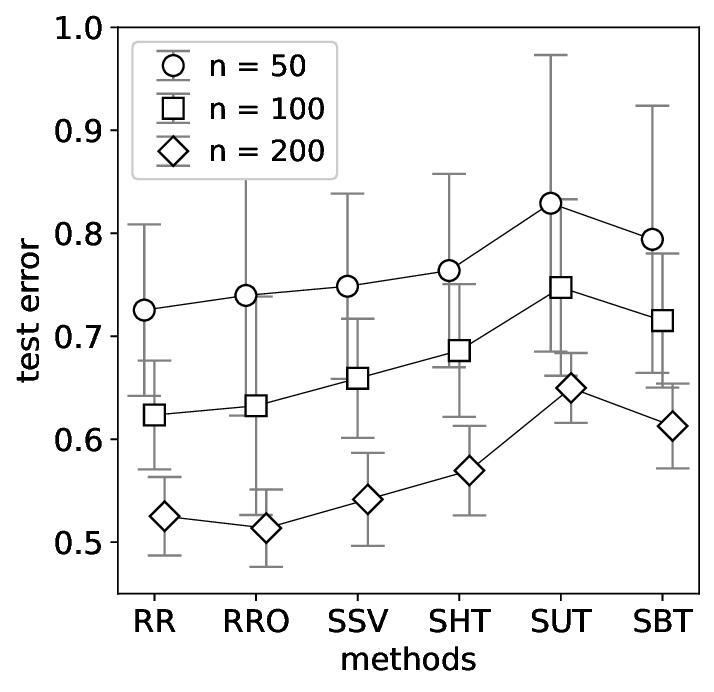}

(b)  Auction dataset
\end{center}
\end{minipage}

\begin{minipage}[t]{\figwidth}
\begin{center}
\includegraphics[width=\figwidth]{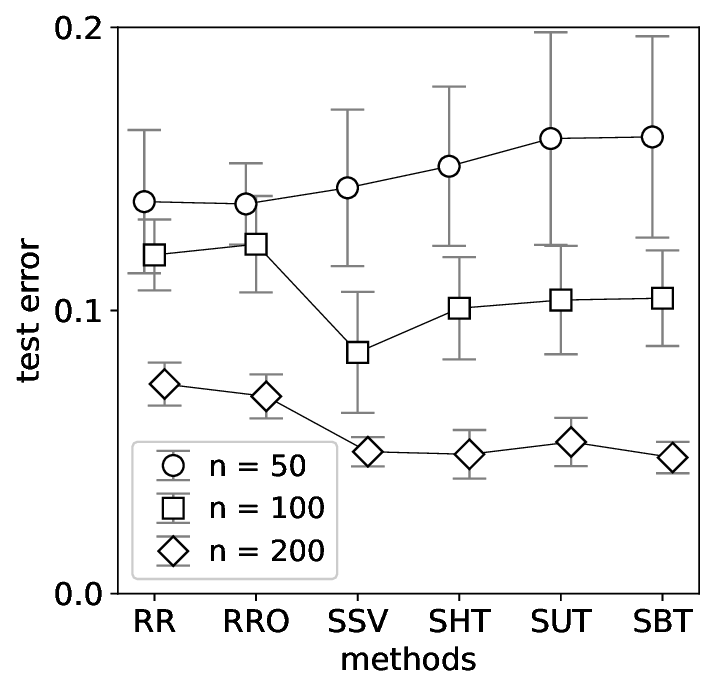}

(c) Energy dataset
\end{center}
\end{minipage}
\begin{minipage}[t]{\figwidth}
\begin{center}
\includegraphics[width=\figwidth]{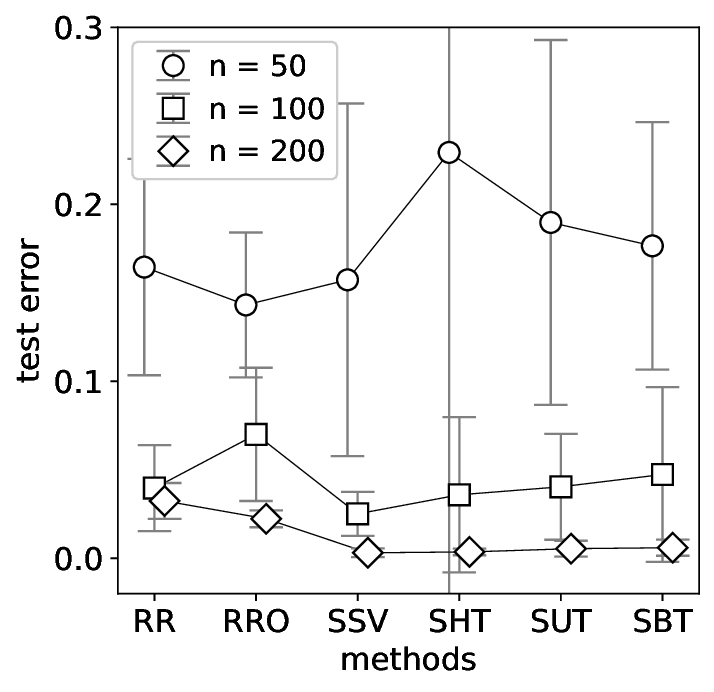}

(d) Yacht dataset
\end{center}
\end{minipage}
\end{center}
\caption{Method comparisons}
\label{fig:methods-comparison}
\end{figure}

On the other hand, as seen in Fig.\ref{fig:methods-comparison} (c) and
(d), for Energy and Yacht datasets, SVD regression methods outperform
ridge regression methods for $n=200$. This advantage of SVD regression
methods is reliable since standard deviations (error bars) are small
enough for $n=200$. This result implies that, for Energy and Yacht
datasets, feature extraction via the SVD is effective for prediction,
which may be the case pointed out in \cite{Hastie+2022,Dhillon+2013}.
Although the performance of SBT is slightly superior to the other SVD
regression methods for Energy dataset, it is not notable and
performances of output-dependent thresholding in SHT, SUT and SBT are
entirely comparable to that of output-independent thresholding in SSV.
Note that, by Fig \ref{fig:methods-comparison}, the difference of
performances between ridge regression methods and SVD regression methods
may not so clear when the number of training data is small.

\begin{figure}[h]
\begin{center}

\begin{minipage}[t]{\figwidth}
\begin{center}
\includegraphics[width=\figwidth]{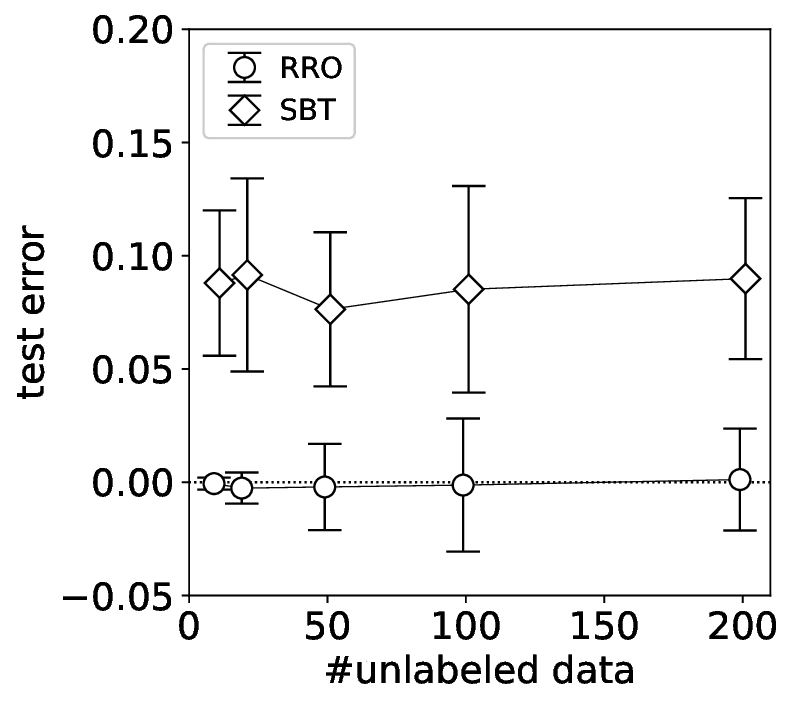}

(a)  GPU dataset
\end{center} 
\end{minipage}
\begin{minipage}[t]{\figwidth}
\begin{center}
\includegraphics[width=\figwidth]{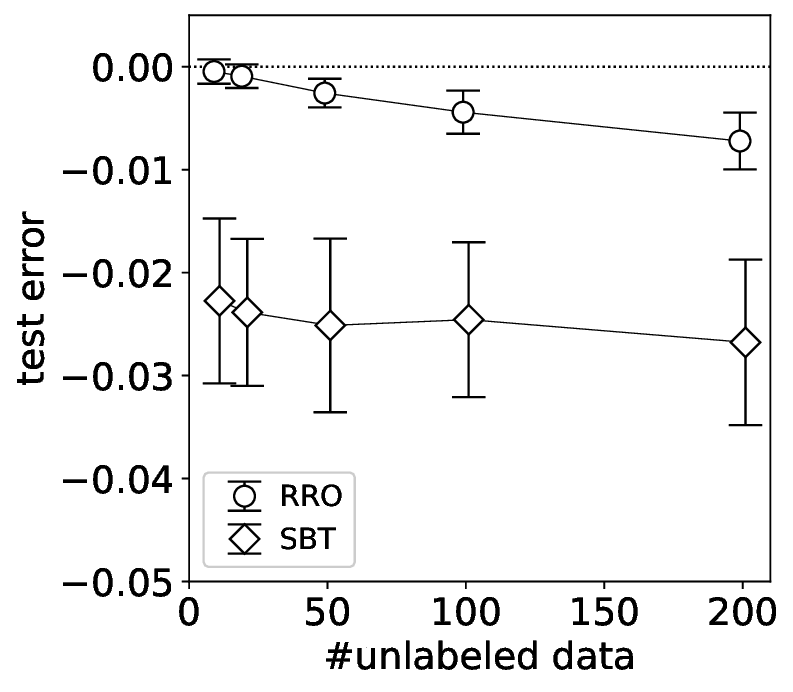}

(b)  Energy dataset
\end{center}
\end{minipage}
\end{center}
\caption{Effect of size of unlabeled samples}
\label{fig:unlabeled-sample-size}
\end{figure}

In the next experiment, we see an effect of the number of unlabeled data
used in the methods. In Fig.\ref{fig:unlabeled-sample-size}, we show the
deviations of test errors of RRO and SBT from test error of RR. We set
$\nunlab=10,20,50,100,200$ under $n=200$. In
Fig.\ref{fig:unlabeled-sample-size} (a) for GPU dataset, for both of RRO
and SBT, the deviations are almost unchanged and those are within the
error bars at any $\nunlab$. This implies that there may be no advantage
of using unlabeled input samples.  In
Fig.\ref{fig:unlabeled-sample-size} (b) for Energy dataset, we can
clearly see the advantage of using unlabeled input samples for
RRO. Also, the average of deviation for SBT tends to decrease while it
is within the error bar. Therefore, the advantage of using SBT may not
so clear. By this result, for semi-supervised learning, our idea of
incorporating unlabeled input samples into a regressor can be effective
depending on datasets and regression methods.

\section{Conclusions and future works}

In this paper, we proposed a SSL scheme using over-parameterized
regression method, in which kernels have information of both of labeled and
unlabeled samples. We then gave several methods based on the MNLS
approach for the over-parameterized regression problem.
We then examined the semi-supervised learning schemes using
over-parameterized regression methods for real data from UCI repository.
The results are summarized in the introduction.

There are several future works. Firstly, the effectiveness of introducing
over-parameterized regression methods into semi-supervised learning is found to
be dataset-dependent. Therefore, we need to clarify when ridge/SVD
regression methods are effective, by which we can choose an appropriate
method depending on datasets.  Moreover, through this investigation, we
need to clarify whether information of unlabeled input samples is helpful
or not in our manner of semi-supervised learning. Next, in SVD
regression methods, output-dependent feature selection schemes via
thresholding methods were not so effective. And, they were worse
depending on datasets.  There may be several reasons which are
computational problem of SVD components with small SVs, accuracy of
variance estimate in SUT and SBT, accuracy of cross validation choice
under SVD decomposition. We need to further investigate and refine the
thresholding methods.

\section*{Acknowledgment}

This work was supported by Japan Society for the Promotion of Science
(JSPS) KAKENHI Grant Number 21K12048. 


\end{document}